\pdfoutput=1

\documentclass[11pt]{article}

\usepackage[final]{coling}

\usepackage{times}
\usepackage{latexsym}
\usepackage{amsmath}
\usepackage{booktabs}
\usepackage{bbm}
\usepackage{cleveref}

\usepackage[T1]{fontenc}

\usepackage[utf8x]{inputenc}

\usepackage{microtype}

\usepackage{inconsolata}

\usepackage{graphicx}
\usepackage{xcolor} 

\usepackage{amsthm} 

\usepackage{listings}
\lstdefinestyle{promptstyle}{
  basicstyle=\ttfamily\small, 
  backgroundcolor=\color{gray!10}, 
  frame=single, 
  rulecolor=\color{gray}, 
  breaklines=true, 
  keywordstyle=\color{blue}\bfseries, 
  stringstyle=\color{red}, 
  numbers=none, 
  captionpos=b, 
  escapeinside={(*@}{@*)}, 
}

\newtheorem{hypothesis}{}

\title{Collective Memory and Narrative Cohesion: A Computational Study of Palestinian Refugee Oral Histories in Lebanon}

\author{Ghadeer Awwad \and Lavinia Dunagan \and David Gamba \and Tamara N. Rayan \\
School of Information \\ University of Michigan  \\ Ann Arbor, MI, USA}

\begin{document}
\maketitle
\begin{abstract}
This study uses the Palestinian Oral History Archive (POHA) to investigate how Palestinian refugee groups in Lebanon sustain a cohesive collective memory of the Nakba through shared narratives. Grounded in Halbwachs’ theory of group memory, we employ statistical analysis of pairwise similarity of narratives, focusing on the influence of shared gender and location. We use textual representation and semantic embeddings of narratives to represent the interviews themselves. Our analysis demonstrates that shared origin is a powerful determinant of narrative similarity across thematic keywords, landmarks, and significant figures, as well as in semantic embeddings of the narratives. Meanwhile, shared residence fosters cohesion, with its impact significantly amplified when paired with shared origin. Additionally, women’s narratives exhibit heightened thematic cohesion, particularly in recounting experiences of the British occupation, underscoring the gendered dimensions of memory formation. This research deepens the understanding of collective memory in diasporic settings, emphasizing the critical role of oral histories in safeguarding Palestinian identity and resisting erasure.
\end{abstract}

\section{Introduction}
The Nakba, a pivotal moment in Palestinian history marked by the mass displacement of approximately 770,000 Palestinians during the forcible establishment of Israel \cite{bisharatDisplacementSocialIdentity1994, khalidiAllThatRemains1992,abu-sittaMappingMyReturn2016, khouryRethinkingNakba2012,khalidiPalestinianIdentity1997,masalhaPalestineNakbaDecolonising2012}, embodies a broader colonial project to establish an ethnocratic state \cite{khouryRethinkingNakba2012, pappeEthnicCleansingPalestine2007, benvenistiIsraelPalestinianRefugees2007, masalhaPalestineNakbaDecolonising2012}. The memory of the Nakba and the immense suffering caused by forced displacement, loss of land, and erasure of the Palestinian presence remains central to the national consciousness of Palestinians \citep{allanRefugeesRevolutionExperiences2013,sadiIntroductionClaimsMemory2007,jayyusiIterabilityCumulativityPresence2007, khalidiPalestinianIdentity1997}. Among Palestinians, refugees in Lebanon are exemplary in their resilience and their maintainance of a collective identity despite being intensely marginalized and disadvantaged. Denied basic rights under Lebanese law, they experience profound spatial, institutional, and economic exclusion \cite{siklawiPalestinianRefugeeCamps2019, rmeileh2021sumud}. Against these hardships, camps have historically served as spaces of resistance, particularly during the 1970s and 1980s when they became bases for the Palestinian Liberation Organization \cite{siklawiPalestinianRefugeeCamps2019, rmeileh2021sumud}. Beyond armed resistance, cultural efforts such as storytelling and poetry are crucial for preserving their national identity and asserting the right to return \cite{siklawiPalestinianRefugeeCamps2019, rmeileh2021sumud, sayighOralHistoryColonialist2015, sayighPalestinianCampWomen1998}.

In consideration of the unique social and political context in which Palestinian refugees in Lebanon are situated, this study is centered on the following research questions: Do Palestinian refugee groups in Lebanon, physically isolated from their homeland, maintain cohesive collective memory formation in their remembrance of the Nakba? And if so, what are the indicators of similarity in their narratives? 

To answer these questions, we employ natural language processing (NLP) techniques in one of the most comprehensive digital repositories of oral histories told by refugees of the Nakba, the Palestinian Oral History Archive (POHA) \cite{sleimanNarratingPalestinePalestinian2018}. POHA has been used extensively to study collective memories and narratives of refugees. We contrast and complement prolific interpretive approaches \citep{davisPalestinianVillageHistories2011, barakatReadingPalestinianAgency2019, swedenburgMemoriesRevolt193619391995,moonOutsideLocusControl2023} that study the sociohistoric impacts of the Nakba and extend other quantitative studies, such as \citet{banatPalestinianRefugeeYouth2018}, which demonstrate the persistence of collective memory in Palestinian youth. Our methodology uses various textual representations of archive interviews that encode themes, specific entities, and semantic content. We then statistically compare pairwise similarities of textual representations of the refugees' narratives. Through this, we measure the degree of similarity within different refugee communities in Lebanon in terms of how members remember the Nakba. As a theoretical framework, we use Halbwachs' \citeyearpar{halbwachsCollectiveMemory1992} concept of group memory to understand how Palestinian remembrance of the Nakba in the diaspora is shaped by boundary-making identity markers, in this case, location and gender. Our main findings show that sharing origin is a strong indicator of higher similarity in refugee narratives, sharing residence in Lebanon is also an indicator of similarity, and sharing both is in many cases an even stronger indicator of cohesion. We also find that the same gender is an indicator of cohesion, although to a smaller degree, as women's narratives are weakly similar to other women. By analyzing the narratives documented within POHA, which represents more than 50 cities and refugee camps in Lebanon, this study contributes understanding of how collective memory formation functions on a larger scale within the Palestinian context.

\section{Background}
\label{sec:background}

Memory work, through oral histories, counters Zionist narratives and challenges the erasure of Palestinian experiences \cite{khouryRethinkingNakba2012,bisharatDisplacementSocialIdentity1994,hanafiPalestinianRefugeesIdentity2011,sayighOralHistoryColonialist2015,masalhaPalestineNakbaDecolonising2012,massadPersistencePalestinianQuestion2006}. By documenting the lived experiences of women, refugees, and everyday Palestinians that are underrepresented in national archival records and historiography \citep{sleimanNarratingPalestinePalestinian2018,allanIntroductionContinuous2021,farahaboubakrPeasantryPalestinianFolktales2017,hanafiPalestinianRefugeesIdentity2011}, oral histories provide a more complex account of the Nakba.  

This complexity is seen in how refugees diverge in how they remember the Nakba. Collective memory formation and positionality often dictate which events, people, and places are remembered based on the concerns that were and continue to be relevant for refugees \citep[p.14]{ben-zeevPalestinianVillageIjzim2002}. As Silmi found in her analysis of oral histories within the Nakba Archive, different “economic, social, and political locations entail different positions, frames of reference and value systems… [such] different positionalities are reflected in the way they remember and tell their stories of life in Palestine and Arab-Jewish relations before the Nakba” \citeyearpar[p.61]{silmiMarginCentreNarrating2021}. Gender produces further multiplicity, seen in Rosemary Sayigh’s \citeyearpar{sayighPalestiniansPeasantsRevolutionaries1979, sayighPalestinianCampWomen1998, sayighProductProducerPalestinian2007} extensive analysis of recorded testimonies in Lebanon’s refugee camps. Sayigh was one of the first to assert that “women are often ‘subversive’ tellers of national history” \citeyearpar[p.47]{sayighPalestinianCampWomen1998} because they include internal conflicts within the resistance movement and detailed accounts of atrocities suppressed within official historiographies. Similarly, Humphries and Khalili \citeyearpar{humphriesGenderNakbaMemory2007} and Khoury \citeyearpar{khouryShufatRefugeeCamp2018} found that Palestinian women choose to transmit information based on what is meaningful within their lived experiences, causing their stories to diverge from "official" histories significantly.

Oral history archives, especially in audiovisual formats, present unique challenges for computational methods due to their non-textual medium \citep{Pessanha_Salah_2021}. \citet{Greenberg_1998} observes that NLP is particularly well-suited for archival collections, as numerous studies have demonstrated its ability to induce new metadata fields and values from archival records or existing metadata. More broadly, NLP can enrich archival practices---it can identify recurring themes, topics, and significant entities such as people, places, or events \citep{Colavizza_Blanke_Jeurgens_Noordegraaf_2021, Jaillant_Rees_2023}. Notably, there is also substantial work in NLP interested in capturing narrative structure \citep{piper-2023-computational, zhuAreNLPModels2023,shuttleworthNarrativesConceptualModels2022, santanaSurveyNarrativeExtraction2023, szojkaNarrativeCoherenceMultiple2020, bendevskiAutomaticNarrativeCoherence2021, akterFaNSFacetbasedNarrative2024}. This line of research at times even pertains to biographical narratives \citep{bamman-smith-2014-unsupervised, brookshire-reiter-2024-modeling}, similar to those in POHA. We largely do not take advantage of these techniques for two reasons---POHA's interviews have been explicitly given structure by archivists and there are few pre-existing resources available for Arabic---but our study speaks to a more collective direction for computational narrative analysis.

Using computational methods, we uncover cohesion in narratives of dispossession and resilience, highlighting how these stories redefine Palestinian identity through collective experiences of loss and resistance \citep{bisharatDisplacementSocialIdentity1994, qumsiyehPopularResistancePalestine2011}. In this paper, we leverage both the metadata included in POHA and transcriptions of interview speech to construct a holistic representation of Nakba narratives. The Nakba-POHA collection thus stands as a testament to the enduring fight for Palestinian rights and dignity and an example of how computational tools can uncover deeper insights from underrepresented histories.

\section{Theoretical Framework and Hypothesis}
\label{sec:theoretical framework and hypothesis}
To understand how Palestinians remember the Nakba, we build upon Halbwachs' \citeyearpar{halbwachsCollectiveMemory1992} theoretical framework of group memory. This theory is appropriate for our analysis of narrative cohesion within refugee communities because it outlines how individual memories are co-constructed through external social relations. Halbwachs observes that individuals will retrieve their memories in alignment with the logic of the group to which they belong. This can involve the reconstruction or rearranging of individual memories so that they have greater coherence with other group members’ memories. In this process, the individual memories of group members will resemble each other and reflect the overall interests and thoughts of the group. However, this phenomenon only holds when the group is strongly established, long-lasting, and able to resist external modes of thinking. If group membership is interrupted through contact with externalities, such as outside beliefs or exposure to a different social context \citep[p.188]{halbwachsCollectiveMemory1992}, the cohesion of recollection can also be interrupted. 

Halbwachs delineates groups as families, religious groups, and social classes that have developed a distinctive unity of outlooks over time \citeyearpar[p.52]{halbwachsCollectiveMemory1992}. Additionally, other potential group formations particular to the Palestinian context, such as the nature of trauma, age group, class, and tribe, could have an impact on narrative cohesion. However we limit our analysis to gender and geography because prior work has proven these aspects to be central identity markers that shape refugees' experiences of community \citeyearpar{masalhaDecolonizingMethodologyReclaiming2018}. As discussed in the background section, Palestinian women exhibit their own gendered memory of the Nakba. Therefore, we situate gender as a boundary-making identity marker. Additionally, geographical displacement from an original location to a foreign location is the key aspect that shapes Palestinian refugee groups and their experience \citep{sayighPalestiniansPeasantsRevolutionaries1979, safranDiasporasModernSocieties1991}. Therefore we situate place of origin and place of residence as another boundary-making identity marker. 

Using this theoretical framework, we test the cohesiveness of the memories of Palestinians that have formed groups within Lebanon. While Palestinian refugees remember the Nakba in a multiplicity of ways, we expect that refugees' boundary-making identity markers will reflect higher cohesion in their narratives.

\begin{hypothesis}
We hypothesize that same interviewee gender would be associated with increased cohesion in narratives (more so than across genders).
\end{hypothesis}

\begin{hypothesis}
We hypothesize that two interviewees belonging to the same spatial group will be associated with increased cohesion in their narratives.
\end{hypothesis}

For our analysis, we define groups as interviewees who have the same gender, place of residence at the time of the interview and/or share the same place of origin from which they were displaced during the Nakba. Cohesion of group memories will be measured in the broader narratives around themes pertaining to the Nakba experience, such as the narration of Zionist invasions and their expulsion, as well as mentions of specific landmarks, significant figures, keywords and places. We expect that when refugees belong to the same group, there will be a higher similarity in how group members remember these entities.

\section{Data \& Methods}

\subsection{The Palestinian Oral History Archive}

POHA is an archive of over a thousand video and audio interviews with Palestinian refugees in Lebanon displaced during the Nakba \citep{sleimanNarratingPalestinePalestinian2018}.\footnote{POHA houses distinct archival collections to preserve Palestinian oral history. While its focus is largely on the Nakba, it also includes interviews documenting folklore and traditional tales; we filtered these out for our analysis.} Although POHA is designed as an audiovisual archive \citep{sleimanNarratingPalestinePalestinian2018}, our approach focuses on the transcripts of the interviews. We opted for textual processing techniques to facilitate medium to large-scale analysis, acknowledging the potential for future multimodal analyses incorporating audiovisual features. Each POHA interview entry also includes metadata about the interviewee(s), interviewer(s), and content. We extracted this information from the POHA website. The content metadata includes keywords, headers summarizing main topics, significant figures, landmarks mentioned, and Table of Contents headers (TOCs) outlining each interview's structure. Details about the origin and residence of interviewees provide valuable context for understanding the narratives (see \cref{sec:appendix:data:extraction} for extraction details). To convert the audio content into text, we used the API of a transcription service, Transkriptor \citep{transkriptor}. We discuss this process in  \cref{sec:appendix:data:transcription}.

\subsection{Measuring Interviewees' Narrative Cohesion}

\begin{table}[]
\centering
\caption{Aspects that we represent in different vectors and the sources for representation, specifying which ones use a BoW representation and which one use a semantic embedding using a LLM.}
\label{tab:representations}
\resizebox{\columnwidth}{!}{%
\begin{tabular}{@{}lll@{}}
\toprule
\textbf{Aspect} & \textbf{Source} & \textbf{Representation} \\ \midrule
Keywords/themes & Metadata & BoW \\
Significant Figures & Metadata & BoW \\
Families & Metadata & BoW \\
Landmarks & Metadata & BoW \\
NER & Transcripts & BoW \\
Semantic by theme & Transcripts & Semantic Emb \\ \bottomrule
\end{tabular}
}
\end{table}

We measure cohesion among interviews by comparing their similarities on different representations. We use various aspects of the interviewee narratives data to construct these representations. Table~\ref{tab:representations} outlines the different aspects we encode into representations and the methods used. We employ two main forms of representation:

\begin{description} 
    \item[Bag-of-Words (BoW):] Applied primarily to metadata to represent specific entities and themes (denoted by curated keywords). 
    \item[Semantic Embeddings:] Dense vector representations obtained from the transcripts using a large language model, interpreted as capturing broader semantic similarity. 
\end{description}

In the following sections, we provide more specifics on these representations.

\subsubsection{Bag-of-Words Representations of Key Entities}

We use bag-of-words (BoW) representations to capture specific entities related to various aspects of the interviews. These are listed in the metadata curated by archivists, focusing on landmarks, families, significant figures, and keywords. The keywords are curated to form a thematic representation of the interview \citep{sleimanNarratingPalestinePalestinian2018}. Additionally, we extract general named entities from the transcripts using an Arabic NER model from the CAMeL Tools library \citep{obeid-etal-2020-camel} (details in \cref{sec:appendix:methods:ner}). These entities capture mentions of places or landmarks without proper names, contributing to a nuanced understanding of similarity. Comparisons using BoW representations emphasize narrative similarity based on shared entities like families or significant figures. In contrast, keyword-based similarity reflects shared broad themes between interviews.

\subsubsection{Semantic Embeddings from Transcripts}
We complement our analysis by measuring narrative similarity through semantic embeddings of the transcripts. Utilizing instruction-conditioned embeddings \citep{su-etal-2023-one} and OpenAI's \texttt{text-embedding-3-large} model, we generated vector representations for interview sections (see \cref{sec:appendix:methods:embeddings}). Notably, we did not filter out interviewers' speech; since we wanted to retain the full narrative structure of each interview as a text, and since \citet{sleimanNarratingPalestinePalestinian2018} describe a high degree of consistency in their approach. This indicates that the interviewers follow the same structure and that the embeddings reflect the interviewees' narratives.\footnote{Any filtering method barring that based on speaker identity could also potentially eliminate valuable testimony from interviewees.}

\begin{figure}
    \centering
    \includegraphics[width=1\linewidth]{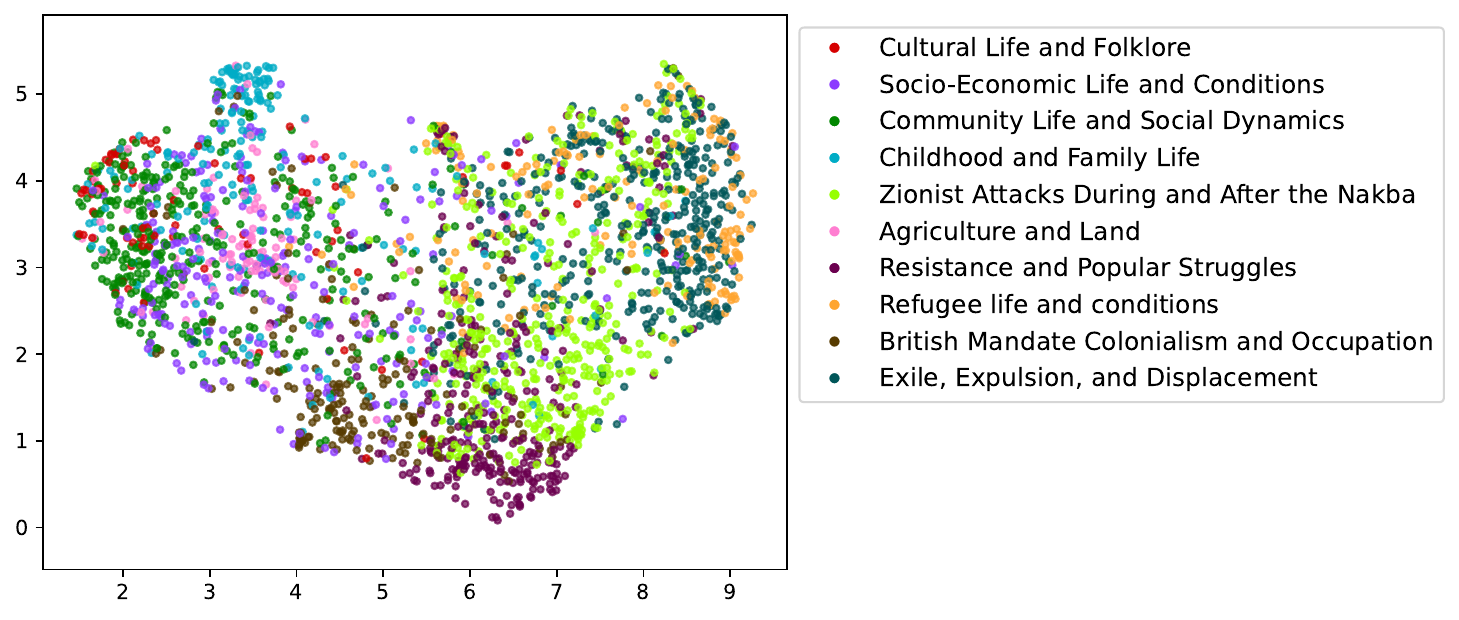}
    \caption{UMAP of instruction-based embeddings of interview transcript sections for all themes. Related themes are visually indistinguishable in this decomposition, while different themes are distant from each other.}
    \label{fig:umap_proj}
\end{figure}

To avoid oversimplifying the diverse interview content, we partitioned transcripts thematically using metadata-derived characteristics. Each interview's archivist-curated Table of Contents (TOC) provides descriptive headers for sections. We perform thematic narrative comparisons, by extracting embeddings from excerpts corresponding to TOC subheaders. We categorized the TOC headers into themes to compare excerpts across interviews (details in \cref{sec:appendix:methods:themes}). The resulting embeddings are visualized using UMAP \citep{McInnes2018} in \cref{fig:umap_proj}. Additional details on obtaining the embeddings are provided in \cref{sec:appendix:methods:embeddings}.

For analysis, we focused on the four more prevalent themes closely related to the Nakba—Zionist Attacks During and After the Nakba; Exile, Expulsion, and Displacement; British Mandate Colonialism and Occupation; and Resistance and Popular Struggles—we created embeddings for transcript excerpts within these themes. 

\subsubsection{Similarity as cohesion}

Given representations $\mathbf{e}^A_i, \mathbf{e}^A_j$ of interviews $i, j$, where the $e^A_i$ could be BoW representations of metadata entities or semantic embeddings of the transcripts, we measure the cohesion of the interviews via the cosine similarity $\sigma(\mathbf{e}^A_j, \mathbf{e}^A_j)$ between the representations

\begin{equation}
    \sigma(\mathbf{e}^A_j, \mathbf{e}^A_j) = \frac{\mathbf{e}^A_j \cdot \mathbf{e}^A_j}{\|\mathbf{e}^A_j\| \|\mathbf{e}^A_j\|}.
\end{equation}

Despite using the same measure for similarity, We heavily lean into the meanings of each of the aspects to interpret the meaning of similarity and cohesion contingent on what is being represented.

\subsection{Statistical Analysis of Narrative Cohesion}
\label{sec:methods:models}

We aim to compare pairs of interviews given interviewee identity markers. For example, we test whether pairs where both interviewees are female are more or less similar than pairs where both are male. To achieve this, we estimate models where the dependent variable is the similarity score, and the independent variables encode identity or community comparisons of interest.

\subsubsection{Analysis by Gender Pairings}
\label{subsec:gender}

To analyze differences by gender, we use the following model with the standarized similarity $\sigma^A_{ij} = \sigma(\mathbf{e}^A_i, \mathbf{e}^A_j)$ as the dependent variable:

\begin{equation}
    \sigma^A_{ij} = \sum_{X \in \{FF, FM,MM\}} \beta_{X} \, g^{X}_{ij} + \mu_{i} + \eta_{j} + \epsilon_{ij}.
\end{equation}

Here, $g^{X}_{ij} = \mathbbm{1}(g_{ij} = X)$ are dummy variables indicating gender pairing ($X \in \{FF,FM,MM\}$). Then, $\beta_{X}$ captures the effects of the gender pairings. Random effects $\mu_i \sim \mathcal{N}(0, \sigma^2_{\mu})$ and $\eta_j \sim \mathcal{N}(0, \sigma^2_{\eta})$ account for interviews $i$ and $j$, respectively, and the residual error is $\epsilon_{ij} \sim \mathcal{N}(0, \sigma^2_{\epsilon})$.

\subsubsection{Analysis by Community(Location) Pairings}
\label{subsec:community}

To compare similarities $\sigma^T$ (per theme) based on location, we use:

\begin{equation}
    \sigma^T_{ij} = \beta_0 + \beta_1 s^o_{ij} + \beta_2 s^r_{ij} + \beta_3 (s^o_{ij} \times s^r_{ij}) + \mu_{i} + \eta_{j} + \epsilon_{ij}.
\end{equation}

In this model, $s^o_{ij}$ and $s^r_{ij}$ are binary indicators of whether interviews $i$ and $j$ share the same origin or residence, respectively. The interaction term captures the combined effect of shared origin and residence. Random effects and residuals are defined as before.

Due to imbalance—with few pairs sharing both origin and residence—we apply an inverse weighting scheme based on the prevalence of each pairing type (details in \cref{sec:appendix:models:weighted}). To better capture uncertainty, we also ran the models within a Bayesian framework using the \texttt{brms} package \citep{burknerBrmsPackageBayesian2017}, allowing us to compare credible intervals with the frequentist results, we detail more in \cref{sec:appendix:models:bayesian}.

\section{Findings}

Using the models outlined, we test cohesion for both gender and location along a set of different aspects. As discussed in Section \ref{sec:theoretical framework and hypothesis}, we hypothesize that narrative cohesion will be higher 1) within same-gender pairs and 2) within same-community pairs (place of origin and place of residence). We refer to analyses of BOW representations of metadata and extracted named entities as relating to ``factual'' aspects of the interviews and the analyses using embeddings of different interview sections as ``thematic.''

\subsection{Women's narratives are slightly more cohesive}
\label{sec:findings:gender}

We begin by testing cohesion within same-gender pairs. Palestinian women testify to different experiences of all three temporalities represented in POHA's biographies (pre-1948 village life, the Nakba, and refugee life); we also assume that group memory is operative on some level. While we do not expect that woman-woman or man-man pairs will necessarily display more or less cohesion according to our measure, we do expect that their levels of cohesion will be separable compared to the standard of man-woman pairs.

\begin{figure}[t!]
    \centering
    \includegraphics[width=\linewidth]{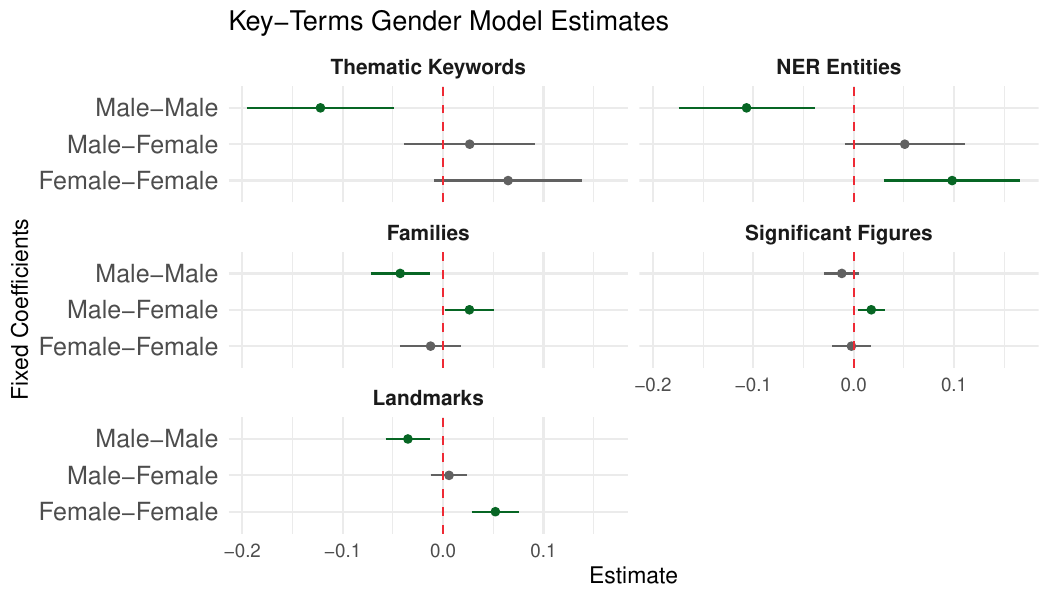}
    \caption{Model coefficients reflecting the relationship between factual mention-based cohesion and gender. Pairs of interviews where both interviewees are men tend to be less similar than those where both interviewees are women or those where one is a man and one is a woman; interviews where both interviewees are women are usually more similar, especially for the less specific measures.}
    \label{fig:kwgender}
\end{figure}

We use the model specified above in \ref{subsec:gender} to iteratively test the relationship between gender and different bag-of-words descriptors of the narrative. This yields the coefficients displayed in Figure \ref{fig:kwgender}. Our results suggest that there is evidence for women displaying thematic narrative cohesion but not necessarily factual cohesion.

\begin{figure}[t!]
    \centering
    \includegraphics[width=\linewidth]{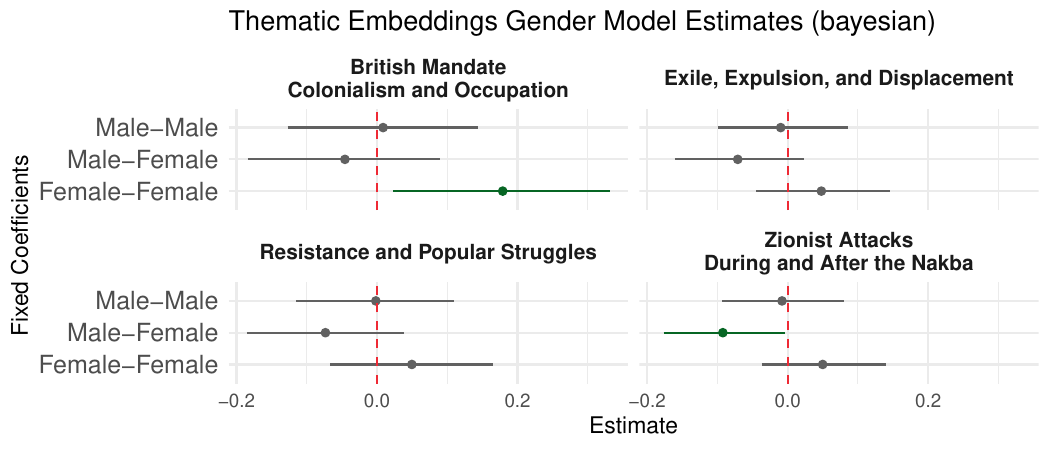}
    \caption{Model coefficients reflecting the relationship between thematic cohesion within topics and gender. Pairs of interviews where both interviewees are men may be slightly more similar than the reference class of man-woman pairs, and pairs of interviewees where both interviewees are women are the most similar of all.}
    \label{fig:tocgender}
\end{figure}

Our original hypothesis about the particularity of women's experiences is only partially supported. In the context of factual mentions, women display only more cohesion than men in the context of landmarks and named entities actually found in the transcript. Otherwise, men seem to diverge in their mentions of landmarks and families as well as keywords and named entities. Thematically, women are observably cohesive only in discussions of the British Mandate. This speaks to how gender is not necessarily primary to the parties and places mentioned in life narratives in POHA.

The specific contexts in which there is observable similarity within gender pairs confirm arguments made in the qualitative literature we inventory in Section \ref{sec:background}. The segments of interviews in which women talk about subjects related to village life in the pre-Nakba period are particulary cohesive; however, women also display significant similarity in discussions of Zionist attacks. In contrast, men appear to have descriptions of both the pre-Nakba period and the post-Nakba period (corresponding to the British Mandate and refugee life categories) that are not notably similar.

\subsection{Shared communities lead to shared narratives}
\label{sec:findings:location}

We also test group memory cohesion in the context of locations. Qualitative work on the experience of Palestinians in Lebanon has observed the development of shared narratives about both refugee life and the Nakba itself. We expect that both interviewees' origins and their current community impact their personal narration of Palestinian history.

\begin{figure}[t!]
    \centering
    \includegraphics[width=\linewidth]{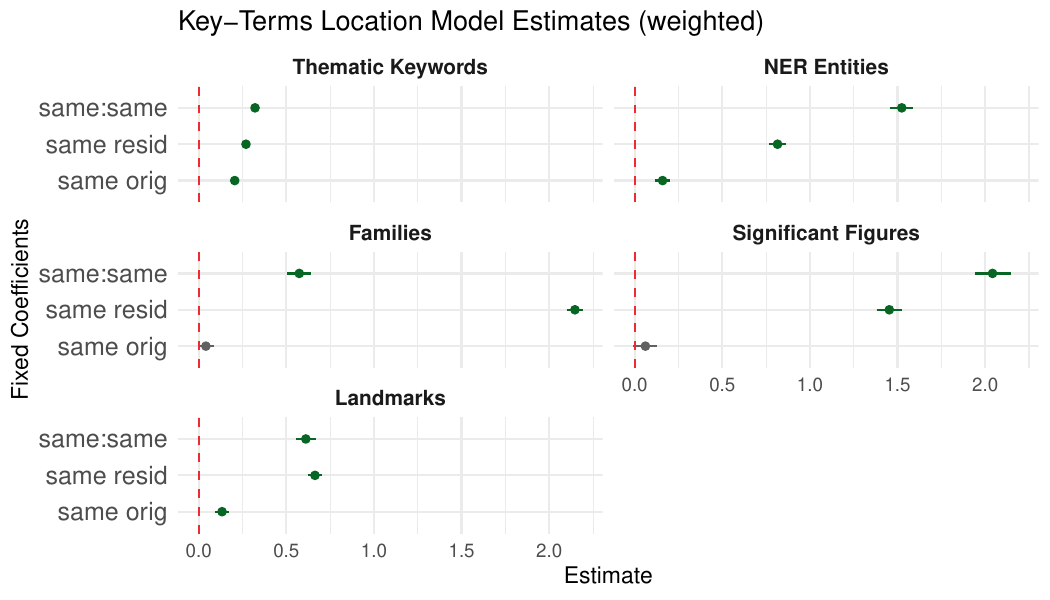}
    \caption{Model coefficients reflecting the relationship between factual mention-based cohesion and shared location. Same origin and resid. consistently predicts the mention of similar entities and (for the keywords) broader concepts. Being from the same place and living in the same place often has a large positive correlation with similarity.}
    \label{fig:kw_location}
\end{figure}

\begin{figure}[t!]
    \centering
    \includegraphics[width=\linewidth]{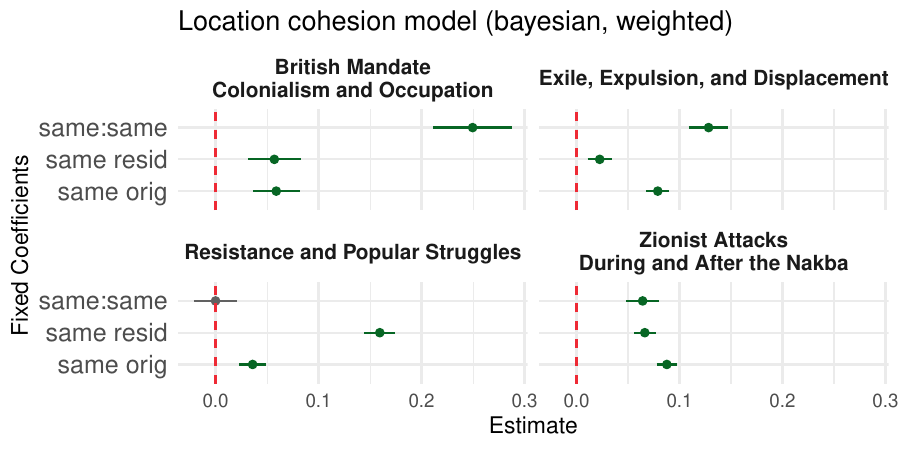}
    \caption{Model coefficients reflecting the relationship between thematic cohesion within topics and shared location. Similar to Figure \ref{fig:kw_location}, being from the same homeplace or residence consistently predicts higher similarities in every major theme. Living in the same place \textit{and} being from the same location is quite predictive of narrating major historical events similarly}
    \label{fig:toc_location}
\end{figure}

We use a mixed effects model, including an interaction term representing the sharing of both original and current residences. This corresponds to an unusual commonality in experiences, given the importance of both interviewees' places of origin and their condition in exile. As shown in Figures \ref{fig:kw_location} and \ref{fig:toc_location}, we find that shared origin \textit{and} shared residence in Lebanon predict shared themes while narrating certain salient aspects of the Nakba.

We find strong support for our original hypothesis. Nevertheless, our results speak to a more nuanced relationship between shared current residence and narrative similarity than one might previously expect. Shared current residence alone predicts similarity along every axis we examine aside from discussions of community life (implicitly in the pre-Nakba context); shared origin alone predicts cohesiveness in both factual mentions and thematic narration, albeit moreso for factual mentions. However, sharing \textit{both} residence \textit{and} place of origin is more predictive than place of origin alone for categories related to both the pre and post-expulsion periods. This suggests processes of collective narration in the refugee context mediate the inclusion of shared themes in individuals' life narratives, depending on whether or not those individuals are from the same place originally.

\section{Discussion}
We observe evidence of group memory in several different scenarios concerning boundary-making identity markers. When interviewees share a community space (same origin or same residence), there is higher similarity within their narratives. Critically, sharing both same origin and residence often results in an even stronger similarity in interviewees' narratives. This indicates that the cohesiveness of narratives depends on the continuity of the group, which, in turn, depends on the continuity of shared space. As well, we observe weaker evidence that gender is an indicator of similarity, though women's narratives tend to be more similar.

The cohesion of narratives based on location can be reasonably explained through the theoretical framework of group memory. Refugees sharing same origin but not the same residence still have cohesive narratives, demonstrating that group members' bond to other members of their original village are not severed in exile. As Toivanen and Baser \citeyearpar{toivanenRememberingDiasporicSpaces2019} state, diasporic memory is not constrained by spatial boundaries, therefore refugees can participate in group recollection with others from the same origins even when they are physically separate. When traumatic experiences threaten group bonds, members will "erase from its memory all that might separate individuals" \citep[p.183]{halbwachsCollectiveMemory1992}, allowing for a collective narrative to emerge from individual memories. Therefore, the cohesion of narratives when refugees share the same origin but not the same residence is evidence of the strength and continuity of their bond despite the violent expulsion of the Nakba. 

Refugees who share the same residence but not the same origin also have cohesive narratives, which shows that the Nakba itself was a catalyst for the formation of new group boundaries. It is anticipated that place of residence impacts group formation because studies \citep{brubakerDiasporaDiaspora2005,safranDiasporasModernSocieties1991,butlerDefiningDiasporaRefining2001} show that different segments of a diaspora will engage in boundary-maintenance and preserve a distinct identity due to self-segregation or social exclusion from the host society. Considering Palestinian refugees in Lebanon continue to be denied citizenship, legal identity, or work visas, their isolation from the rest of Lebanese society has caused refugees to turn inwards and develop strong group bonds with each other. Sharing the same place of residence, even without a continous bond stretching back to the homeland, therefore contributes to preserving a shared identity and cohesive memory as a group. 

Finally, sharing the same origin and place of residence is often a strong indicator of similarity. This is consistent with the framework of group memory, which states that groups maintain cohesive memories so long as they are resistant to outside influences \citep{halbwachsCollectiveMemory1992}. When Palestinians from the same village were forcibly displaced to the same place in Lebanon, their group continuity was maintained. This is not the case for all memories, however, as refugees in the same residence narrativize refugee life and conditions dissimilarly. These memories are newer compared to those from the Nakba period, so it is possible this specific theme has not yet been subjected to the processes of group memory. Overall, however, existing research states that diasporic group memory is shaped by both the homeland and the host country \citep{toivanenRememberingDiasporicSpaces2019, voytivDiasporicGroupBoundaries2024}, therefore it is not unexpected that Palestinian refugees who share the same past and present location have an even stronger similarity in their narratives than refugees who ohare only one location. 

Our findings regarding gender merit further discussion. Gender narrative similarity is more variable and there is no single identifier of a stronger similarity or dissimilarity. This speaks to an overall shared understanding of the Nakba transcending gender boundaries. However, we highlight significant differences. Namely, increased cohesion concerning themes in general (keywords) and pre-Nakba village context. This connects to two factors that influence group memory. First, the constructed social and cultural roles of gender among Palestinians in refugee camps, and second, the nature of refugee camps as spaces of displacement and vulnerability, which operate under unique socio-economic constraints that profoundly shape gender dynamics. Second, pre-Nakba, villages were closely knit communities where women's roles and responsibilities were shaped by communal practices and survival strategies, including supporting resistance \cite{sayighProductProducerPalestinian2007, tamariMountainSeaEssays2008, nimrFastForwardLook2008, yahyaOralHistoryDual2017}. In these settings, women often internalized dominant national narratives shaped by the era, which laid the foundation for their role in the broader narrative of resistance \cite{tamariMountainSeaEssays2008, sayighPalestinianCampWomen1998, yahyaOralHistoryDual2017}.

Women’s factual cohesion appears strongest when referencing specific landmarks or entities in transcripts. This can be associated with their emotional connection to physical places. In contrast, men’s narratives exhibit notable divergence in their descriptions of landmarks, families, and other entities. This suggests that men's experiences are either more individualized or less influenced by communal memory practices. Sayigh’s ethnography corroborates this, highlighting men’s fragmented recollections shaped by personal trajectories and interactions with political and economic systems \cite{sayighPalestinianCampWomen1998, sayighOralHistoryColonialist2015}.

Overall, our findings challenge the notion that same-gender pairs inherently produce cohesive narratives \cite{tamariMountainSeaEssays2008, nimrFastForwardLook2008, sayighOralHistoryColonialist2015, sayighPalestinianCampWomen1998}. Instead, cohesion reflects the collective memory fostered by women’s roles in village life and under colonial rule, even as factual details vary. 

\section{Conclusion}

This paper constitutes a large-scale computational analysis of an oral history archive of the Nakba using natural language processing. Drawing from qualitative literature on Palestinian refugees' experiences, we explore how boundary-making identity markers shape their life narratives. Comparing the narratives of Palestinians beyond Lebanon or the narratives of those from rural vs. urban contexts are opportunities for further research into how locality shapes narrative cohesion. This study, however, is bound by the scope of POHA which focuses solely on Lebanon and offers a preliminary foray into one means of capturing multidimensional narrative cohesion in an archive. With these findings, we hope to both reaffirm and extend prior scholarship on Palestinian collective memory, by providing empirical evidence of how oral narratives allow refugees to build continuity and counteract the violent fragmentation of Palestinian society after the Nakba.

\section*{Limitations}

From a substantive standpoint, the analysis and interviews (which represent just a fraction of the Palestinian refugee population in Lebanon) provide only a partial perspective on the Nakba. Capturing the depth of this historical tragedy in a single analysis is inherently difficult, especially when the Nakba remains an ongoing condition of exile.

Our analysis uses mixed effects models to explore the relationship between cohesion and community membership; here, we identify assumptions that may not be robust. Although we tried a triangulation approach with different aspects, the textual representations are limited in capturing narrative structure. Generally, our focused approach only begins to characterize POHA. Future research could explore other methods for understanding oral history archives, complementing interpretive approaches that engage directly with individual interviews. In addition, the metadata results in this paper partly reflect curation decisions by the POHA team. Archivists' positionality often influences descriptive practices \citep{kingArchivalMetametadataRevision2024,carbajalCriticalDigitalArchives2021}, introducing potential bias in metadata terminology. Discrete metadata fields also risk reducing the rich lived experiences in interviews to simplified categories, although we use semantic embeddings to counteract these limitations. 

\section*{Acknowledgments}
We deeply honor the courage and resilience of Palestinians who have shared their powerful and often overwhelming experiences. Their stories are a testament to the unyielding spirit of resistance, serving not only as a bridge to future generations but as a vital contribution to the ongoing journey toward justice and liberation. We are also grateful to the American University of Beirut librarians for their support in addressing our inquiries.

\bibliography{research-pal,missing_sources}

\appendix
\section{Data Appendix}
\label{sec:appendix:data}

\subsection{Extraction of POHA data}
\label{sec:appendix:data:extraction}

We extracted both audio files and corresponding metadata from the Palestinian Oral History Archive by scraping their website, located on the American University of Beirut's website at \href{https://libraries.aub.edu.lb/poha/}{libraries.aub.edu.lb/poha/}. Every record in POHA has its own dedicated ID, running from \texttt{4158} to \texttt{4887}; each record's page has the URL \texttt{https://libraries.aub.edu.lb/poha/Record/\{ID\}}.

For each interview page, we use the \texttt{BeautifulSoup} library \citep{bs4doc} to extract data from the HTML. Some of the fields (e.g., significant figures who appear in an interview) are variously present and absent; we flexibly add these to JSON representations of the metadata. Interview audio was downloaded from the ``Interview Audio/Video'' field present for most interviews. A minority of the interviews did not have available recordings; after this scraping process, during which we excluded those, we were left with 720 interviews.

\subsection{Extracting Information from Bios}

Our initial extracted metadata directly from POHA did not include specific fields for gender or place of residence. However, this information was sometimes present also within the metadata, just not in a structured format. When available, such information was often embedded within the \textit{bio} field.

To extract gender and place of residence from the bios—which was essential for our study—we utilized a large language model. We compiled the bios in both Arabic and English and supplied them to \texttt{gpt-4o} \cite{openai2024gpt4technicalreport} using the following prompt:

\begin{lstlisting}[style=promptstyle, caption={Prompt to extract structured information from the bios}]
From the following *interview description* + *metadata on names and places of origin* extract the following information in JSON format.

The input data has the following structure.

{{
  "en": {{"bio": "<bio in english>",
          interviewee: ["<name person 1>", "<name person 2>", ...],
          place_of_origin: ["<origin person 1>", "<origin person 2>", ...]}}
  "ar": {{"bio": "<bio in arabic>",
          interviewee: ["<name person 1>", "<name person 2>", ...],
          place_of_origin: ["<origin person 1>", "<origin person 2>", ...]}},
}}


The output data must have the following structure. Not all fields might be present, in that case, omit those

```json
{{
  "en": {{  # for english data extraction
    "interview_date": ""
    "interview_location": ""
    "interviewees": [
      {{
        "name": "",
        "place_of_origin": "",
        "gender": "",
        "birth_date": "",
        "place_of_residence": "",
        "occupation": "",
        "occupation_entity": ""
      }},
    ]
  }},
  "ar": {{  # for arabic data extraction
    "interview_date": ""
    "interview_location": ""
    "interviewees": [
      {{
        "name": "",
        "place_of_origin": "",
        "gender": "",
        "birth_date": "",
        "place_of_residence": "",
        "occupation": "",
        "occupation_entity": ""
      }},
    ]
  }}
}}
```

Notes:
- The interview might have more than one interviewee.
- The description is given in english and in arabic, extract the data for both versions, make sure the order of interviewees is the same for both versions
- Output the `name` as it appears in `interviewee` list. If the name does not appear in the interview list use the name as it appears on the bio
- Output the `place_of_origin` as it appears in `place_or_origin` list. If the place does not appear in the place_of_origin list use the place as it appears on the bio
- Use date format %Y-%m-%d (eg. 2024-08-25) if possible, if not, use %Y (eg. 2024)
- Be careful with keeping the correct transliteration of arabic names and places

----------------------
Interview description:

{{
  "en": {{"bio": "{data["en_bio"]}",
          interviewee: {data["en_interviewee"]},
          place_of_origin: {data["en_place_of_origin"]}}}
  "ar": {{"bio": "{data["ar_bio"]}",
          interviewee: {data["ar_interviewee"]},
          place_of_origin: {data["ar_place_of_origin"]}}}
}}
\end{lstlisting}

Note that in addition to the bios, we provided the interviewees' names and places of origin as recorded in the metadata. This ensured that the model used consistent names, simplifying the process of linking all data together.

\subsection{Curating Places of Residence}

Initially, our metadata extraction did not include direct information about the place of residence for each interviewee. While the bios sometimes contained additional details beyond the specified metadata—including place of residence—this was not always the case. As explained in the previous section, we used \texttt{gpt-4o} to extract structured data from the bios.

After this process, many interviews still lacked specified places of residence in the bios, necessitating an alternative extraction method.

In our second approach, we focused on interviews for which no metadata was available. Arabic-speaking team members manually read the interview transcripts to identify the interviewees' places of residence. This task was inherently time-consuming, requiring approximately 30 minutes per interview for a native speaker due to the length of the interviews and the challenge of obtaining full context.

To expedite this process, we utilized \texttt{gpt-4o} with the following prompt:

\begin{lstlisting}[style=promptstyle, caption={Prompt to Extract Place of Residence from Interview Excerpt}]
From the following excerpt of an interview, extract the place of residence of the interviewee if such place is mentioned.

Notes:
- Do not confuse the place of residence with the place of origin of the interviewee.
- Attempt to identify the place where the interviewee is living at the time of the interview.
- It is very unlikely that the place of residence is within Palestine, as these are stories of exiled refugees.

Respond with the place of residence of the interviewee, the timestamp, and the line of text where you found it.
Also, add a short explanation of why the place is likely to be the place of residence.
If no place of residence can be reasonably identified, add a short explanation of why.

Text:
--------
speaker_id; timestamp; text
{text here}
\end{lstlisting}

Here, the text is formatted as specified in the prompt: \verb|speaker_id; timestamp; text|.

Given that the interviews often exceeded the model's context window, we partitioned them by their table of contents and submitted each section individually. After processing a few interviews, our team members gained insights into which sections were more likely to mention the place of residence. Combining these insights with the model's suggestions, we efficiently processed approximately 250 interviews to obtain places of residence.

Out of the interviews that initially lacked residence information, we successfully identified the place of residence for 207 interviews. This left us with 52 interviews where the place of residence remained unidentified or unclear.

\subsection{Deduplication of Places}

We frequently encountered variations in how places of origin and places of residence were referenced within the metadata and bios. For example, a single refugee camp might be referred to in up to eight different ways due to the addition or omission of words, or the inclusion of the area or country name. This inconsistency posed challenges for grouping and matching interviews based on places of origin and residence.

To resolve this issue, we performed a deduplication process using the \texttt{RecordLinker} library \citep{farleyBayesianApproachLinking2020}. This tool allowed us to match locations by comparing both the Arabic and English versions of the names, utilizing edit distance metrics for each language. The process clustered records that likely referred to the same entity. We chose the most prevalent name in the data as the representative name for each cluster, effectively minimizing the impact of misspellings and variations by standardizing the names across the dataset.

After this process, we reduced the number of unique places to approximately 150 different places of origin and 100 different places of residence. We then manually curated these lists to finalize the mappings, ensuring accurate and consistent duplication of the interview information.

\subsection{Transcriptions and validation}
\label{sec:appendix:data:transcription}

Filtering out interviews about folktales, we are left with 724 interviews. This still comprises about a thousand hours of audio and video. We use Transkriptor, a commercial transcription tool, to transcribe the interviews \citep{transkriptor}. Transcriptor supports transcription optimized for specific dialects of Arabic; this is important for the fidelity of our transcriptions of POHA, given that many of the interviews are in Levantine Arabic rather than MSA. While we explored using smaller Whisper \citep{whisper_paper} models, the large volume of interviews and specificity of the data led to low-quality and slowly obtained results.

Two Arabic-speaking members of our team conducted a qualitative assessment of the interviews to validate transcriptions. This process involved manually reviewing a subset of the transcriptions to ensure accuracy and consistency, particularly for dialect-specific nuances.

The translator effectively captured the local Palestinian dialect without censoring words or omitting context, achieving 90 percent accuracy. However, challenges arose from factors such as the quality of the interviews and their structural design. In some cases, the transcription tool struggled to recognize sentence boundaries and properly connect the text, requiring careful contextual interpretation without re-listening to the audio. Additionally, there were minor issues with reading the local dialect, as the transcription tool spelled words differently than how native speakers would naturally express them.

\section{Methods Appendix}

\subsection{Extracting named entities}
\label{sec:appendix:methods:ner}

In order to extract named entities, we use the CAMeL Tools library \citep{obeid-etal-2020-camel}. In particular, we used the \texttt{CAMeL-Lab}
\texttt{/bert-base-arabic-camelbert-msa-ner} model \href{https://huggingface.co/CAMeL-Lab/bert-base-arabic-camelbert-msa-ner}{available on HuggingFace} to conduct flat token classification. \citet{inoue-etal-2021-interplay} describes this model as having a F1 score of 74.1 on dialectal Arabic (under which Levantine Arabic falls), which is sufficient for our purposes given the complementary nature of our analyses.

\subsection{Defining the Themes}
\label{sec:appendix:methods:themes}

Overall, the Table of Contents (TOC) headers contain over 3,000 entries. Unlike keywords and other entities in the metadata—which required selection from a specific list of entities \cite{sleimanNarratingPalestinePalestinian2018}—the TOC headers were largely left to the discretion of the archivists. Each interview is unique; the flexibility of the headers allows for a broad characterization of an interview's content without imposing a strict classification system. This means that although many headers are very similar, there is no unique categorization of TOCs.

To generate the themes, we first took a random sample of the headers in English and then prompted \texttt{gpt-4o} \cite{Hello_GPT_4o} to generate a list of potential themes to cluster the exemplars. This followed an approach similar to QAEmb, where a large language model is prompted to generate a rubric \cite{benaraCraftingInterpretableEmbeddings2024}. We used this as a starting point for manual curation, adding and modifying themes and relocating exemplars as needed.

In the end, we arrived at the following themes:

\begin{description}
    \item[Childhood and Family Life] Experiences and memories of childhood and family life, including upbringing, education, community interactions, and daily life before displacement.
    
    \item[Education and Intellectual Life] Educational experiences, curricula, teaching methods, and intellectual pursuits both within Palestine before 1948 and among the Palestinian diaspora.
    
    \item[Community Life and Social Dynamics] Community structures, social customs, celebrations, and dynamics within Palestinian towns. Mostly represents community pre-Nakba.
    
    \item[Zionist Attacks During and After the Nakba] Zionist attacks, occupation, and conflicts; during and after the Nakba, including invasions, occupations, massacres, battles, and their effects on Palestinians.
    
    \item[Exile, Expulsion, and Displacement] Accounts of expulsion, exile, and displacement during the Nakba, detailing the journeys, hardships, and experiences of leaving Palestine.
    
    \item[Socio-Economic Life and Conditions] Examinations of socio-economic conditions before the Nakba, including employment, agriculture, economic activities, social relations, and health care in Palestinian communities.
    
    \item[Cultural Life and Folklore] Explorations of Palestinian cultural life and folklore, including traditional songs, customs, and stories.
    
    \item[British Mandate Colonialism and Occupation] Discussions of the British Mandate period (1920–1948), focusing on British administration, policies, the role in Zionist colonization, and Palestinian resistance prior to the Nakba.
    
    \item[Resistance and Popular Struggles] Accounts of Palestinian resistance movements and popular struggles against colonialism and occupation.
    
    \item[Women's Roles] Exploration of women's roles in society, including contributions to community rebuilding, activism, rights, and political involvement during displacement.
    
    \item[Agriculture and Land] Descriptions of agricultural practices, land use, and the importance of agriculture to community life in Palestine before the Nakba.
    
    \item[Reflections and Final Thoughts] Personal reflections on the refugee experience, aspirations for the future, attempts to return, and thoughts on migration and displacement.
    
    \item[Refugee Life and Conditions] Examinations of life as refugees, focusing on living conditions, education, aid, social conditions in exile, and discussions on the right of return.
\end{description}

We then classified all TOCs using a multilabel approach by manually annotating a sample of 340 headers. For classification features, we embedded the headers using instructed embeddings with the instruction: ``Represent the Interview Section Header for classifying its main theme.'' We used a test split and compared the performance of different multi-label models, with a one-vs-rest Support Vector Machine Classifier (SVC) offering the best performance with an AUC of 0.98. We also conducted a qualitative spot-check of the results to verify that the assigned themes were appropriate.

\cref{tab:theme-counts} contains the list of curated themes and the associated number of interview TOCs that were classified as belonging to the theme.

\begin{table*}[]
\centering
\caption{Themes and count of interviews headers with such theme, roughly corresponds to the number of interviews that contain the theme. In bold we note the themes that we focus on for analysis because they are both more related to the Nakba topically and are prevalent as well}
\label{tab:theme-counts}
\resizebox{\columnwidth}{!}{%
\begin{tabular}{@{}lr@{}}
\toprule
\textbf{Theme} & \textbf{TOCs count} \\
\midrule
\textbf{Zionist Attacks During and After the Nakba} & 714 \\
\textbf{Community Life and Social Dynamics} & 697 \\
Cultural Life and Folklore & 580 \\
\textbf{Exile, Expulsion, and Displacement} & 548 \\
\textbf{Resistance and Popular Struggles} & 505 \\
\textbf{Socio-Economic Life and Conditions} & 327 \\
\textbf{British Mandate Colonialism and Occupation} & 259 \\
\textbf{Refugee life and conditions} & 255 \\
\textbf{Childhood and Family Life} & 233 \\
Agriculture and Land & 140 \\
Reflections and Final Thoughts & 51 \\
Women's Roles & 40 \\
Education and Intellectual Life & 27 \\
\bottomrule
\end{tabular}%
}
\end{table*}

\subsection{Thematic Embeddings}
\label{sec:appendix:methods:embeddings}
To capture additional semantic similarities beyond what could be established from the POHA metadata, we utilized embeddings generated by Large Language Models (LLMs). Specifically, we employed OpenAI's \texttt{text-embedding-ada-002} model.

Given that individual sections specified in the Table of Contents (TOC) headers could fall into multiple themes, we designed our approach to focus the embeddings on specific themes. This was crucial because themes in personal narratives often intersect; for instance, discussions of expulsions are frequently preceded by accounts of Zionist attacks.

To ensure the embeddings were theme-specific, we constructed prompts by prepending a thematic instruction to the text. The instructions were provided in both English and Arabic to avoid potential biases arising from language differences. Experiments showed no significant differences between the two languages; however, we proceeded with the Arabic prompt for consistency. The instructions used were:

\begin{lstlisting}[style=promptstyle, caption={Instruction prepended to generate embeddings (in English)}]
Represent the following interview transcript for analyzing the theme
\end{lstlisting}


A key challenge was generating a single representation per theme in an interview. Often, multiple headers pertained to the same theme, and individual sections could exceed the model's maximum context window of 8,091 tokens. To address this, we divided the interview texts into manageable chunks, each prepended with the instruction. We obtained embeddings for each chunk and then combined them using max-pooling to create a single representation for the theme within the interview.

We performed validation procedures to ensure the quality of the embeddings. For instance, we verified that embeddings for the same theme were, on average, more similar to each other than to embeddings from different themes—a consistency that was consistently observed. The cosine similarities ranged from approximately 0.30 to 0.80, with the average often exceeding 0.60.

These thematic embeddings allowed us to capture nuanced semantic relationships within the interviews, facilitating more sophisticated analyses of the narratives.

\subsection{Additional model details}
\subsubsection{Weighted Location Models}
\label{sec:appendix:models:weighted}
In our analysis of pairwise comparisons of interview similarity based on the same place of origin and the same place of residence, we identified a significant imbalance in the data. Most interview pairs did not share either place of origin or residence, and less than 1\% of the data involved pairs sharing both place of residence and place of origin. \cref{tab:appendix:models:weights-table} details the counts for each condition comparing same origin and same residence. Addressing this imbalance was crucial, especially since we aimed to use this data for the analysis of themes, where the data would be further subdivided.

To mitigate the effects of this imbalance, we incorporated weights into the regression estimations of the mixed models. Specifically, we used an \textit{inverse frequency weighting scheme}, where each observation is weighted inversely proportional to its frequency in the data. This means that rarer combinations (e.g., interview pairs sharing both place of origin and residence) receive higher weights, ensuring they have a proportionate influence on the model despite their low occurrence.

We applied the same weights, defined by the general prevalence in our sample, to compute models for both Bag-of-Words (BoW) based representations—which use all interviews—and for themes, which involve pairings of interviews matching on the same theme. This approach ensured consistency in the weights applied to the observations across different analyses.

\begin{table}[]
\centering
\caption{Prevalence of conditions comparing same origin and residence among pairs of interviews in the data.}
\label{tab:appendix:models:weights-table}
\resizebox{0.6\columnwidth}{!}{%
\begin{tabular}{@{}lr@{}}
\toprule
\textbf{Location Pair Condition} & \textbf{Pairs} \\ \midrule
Same Origin\\+Same Residence & 251036 \\
Same Origin\\+Diff. Residence & 18990 \\
Diff. Origin\\+Same Residence & 3338 \\
Diff Origin\\+Diff Residence & 634 \\ \bottomrule
\end{tabular}%
}
\end{table}

\subsection{Bayesian Models}
\label{sec:appendix:models:bayesian}
To complement the frequentist mixed models and to better understand the potential behavior of the parameters, we also implemented the models using a Bayesian framework. We estimated the parameters using Markov Chain Monte Carlo methods provided by the \texttt{brms} library \citep{burknerBrmsPackageBayesian2017}, which utilizes the \texttt{Stan} \citep{standevelopmentteamRStanInterfaceStan2024} language and sampler to implement Bayesian multilevel models similar to the \texttt{lmer} models.

We chose Bayesian models because they allow for the estimation of credible intervals for the parameters, providing a probabilistic interpretation of the parameter estimates. This enabled us to confirm our results on the weak association of the parameters and to understand the variability in our estimates more clearly.

The models followed almost identical specifications to those described in \cref{sec:methods:models}, both mathematically and in the code implementation. We delineate them here following.

To compare similarities $\sigma^T_{ij}$ (per theme) based on location within a Bayesian framework, we use the model

\begin{align}
\sigma^T_{ij} &\sim \mathcal{N}(\mu_{ij}, \sigma^2), \\
\mu_{ij} &= \beta_0 + \beta_1 s^o_{ij} + \beta_2 s^r_{ij} + \beta_3 (s^o_{ij} \times s^r_{ij}) \nonumber \\ &+ u_i + v_j,
\end{align}

where $u_i$ and $v_j$ are random effects associated with interviews $i$ and $j$, respectively, accounting for unobserved heterogeneity. $\sigma^2$ is the residual variance (not similarity).

We specify the following prior distributions for the parameters:

\begin{align*}
\beta_0 &\sim \text{Normal}(0, 0.1), \\
\beta_k &\sim \text{Normal}(0, 1), \quad \text{for } k = 1, 2, 3, \\
u_i &\sim \text{Normal}(0, \sigma_u^2), \quad \sigma_u \sim \text{Exponential}(1), \\
v_j &\sim \text{Normal}(0, \sigma_v^2), \quad \sigma_v \sim \text{Exponential}(1), \\
\sigma &\sim \text{Exponential}(1).
\end{align*}

In this Bayesian model:

\begin{itemize}
    \item $\beta_0$ is the intercept with a prior centered at 0 and a small variance, reflecting our initial belief about the central tendency of similarities.
    \item $\beta_1$, $\beta_2$, and $\beta_3$ are the coefficients for the fixed effects with priors reflecting moderate uncertainty.
    \item $\sigma_u$ and $\sigma_v$ are the standard deviations of the random effects, modeled with Exponential priors to ensure positivity and to express a preference for smaller values.
    \item The residual standard deviation $\sigma$ also follows an Exponential prior, promoting regularization.
\end{itemize}

The gender model follows a very similar definition but uses the parameters as described in \cref{sec:methods:models}

We show the results of the Bayesian models in section \cref{sec:appendix:results:bayesian}

\section{Results Appendix}

\subsection{Model results for all themes}
\begin{figure}[t]
    \centering
    \includegraphics[width=\linewidth]{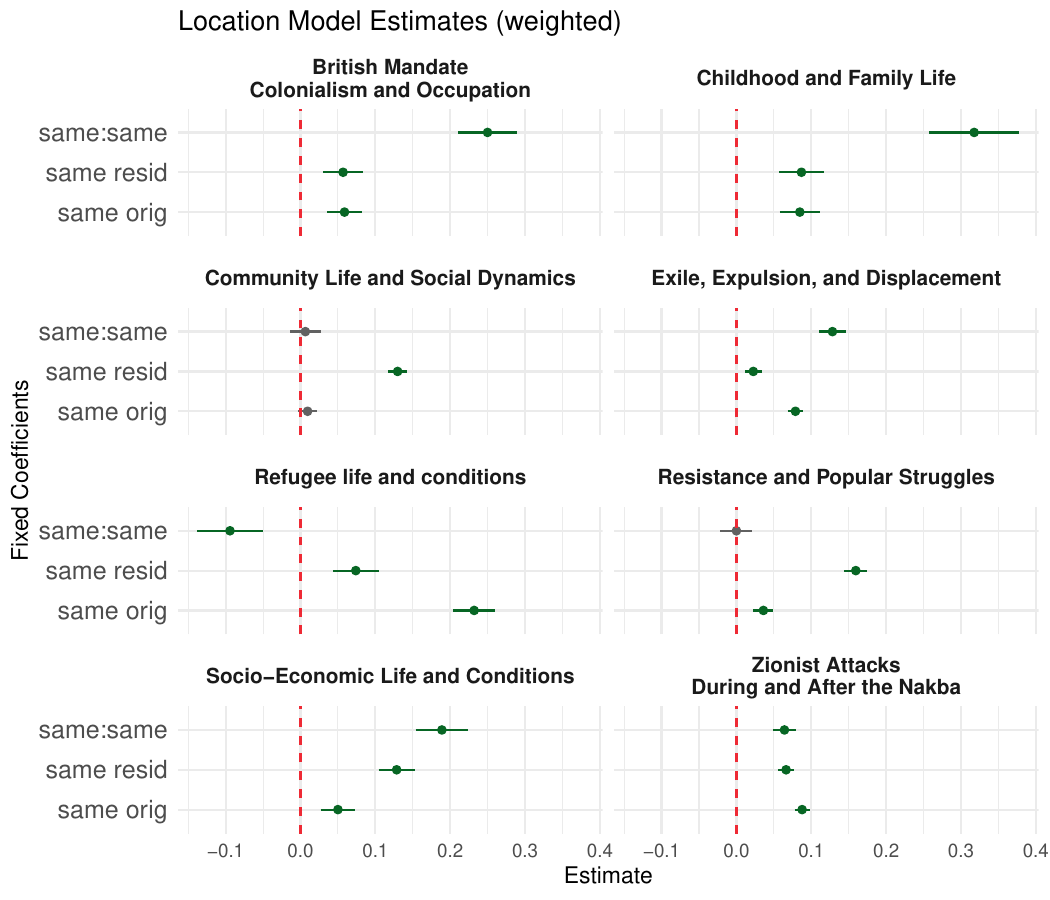}
    \caption{Mixed model for location estimates for all themes, following the same description as \cref{sec:findings:gender}.}
    \label{fig:appendix:results:tocs_location_all}
\end{figure}

\begin{figure}[t]
    \centering
    \includegraphics[width=\linewidth]{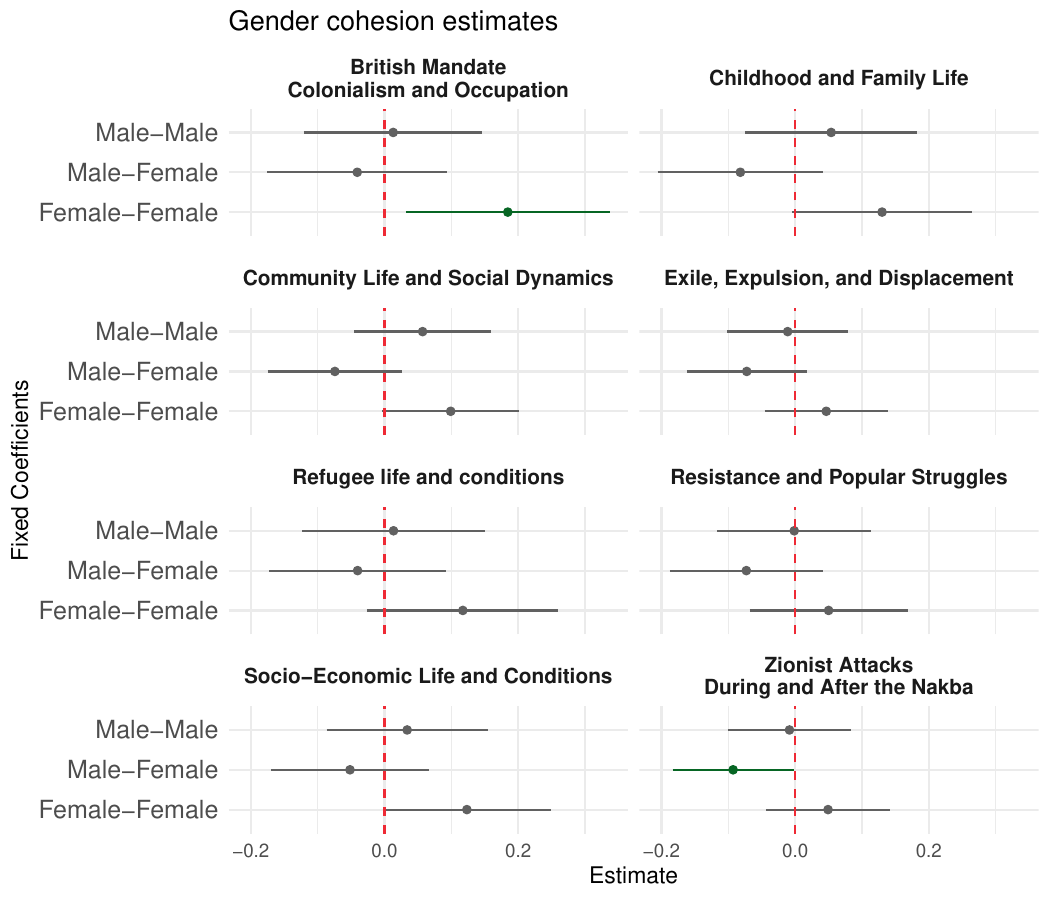}
    \caption{Mixed model for gender estimates for all themes, following the same description as \cref{sec:findings:gender}.}
    \label{fig:appendix:results:tocs_gender_all}
\end{figure}

In the main body, we selected the results for the key 4 themes as being more significant for our analysis perspective on the Nakba. Here we add the thematic plots of all themes, both for location (\cref{fig:appendix:results:tocs_location_all}) and gender (\cref{fig:appendix:results:tocs_gender_all}).

\subsection{Bayesian model results}
\label{sec:appendix:results:bayesian}

\begin{figure}[t]
    \centering
    \includegraphics[width=\linewidth]{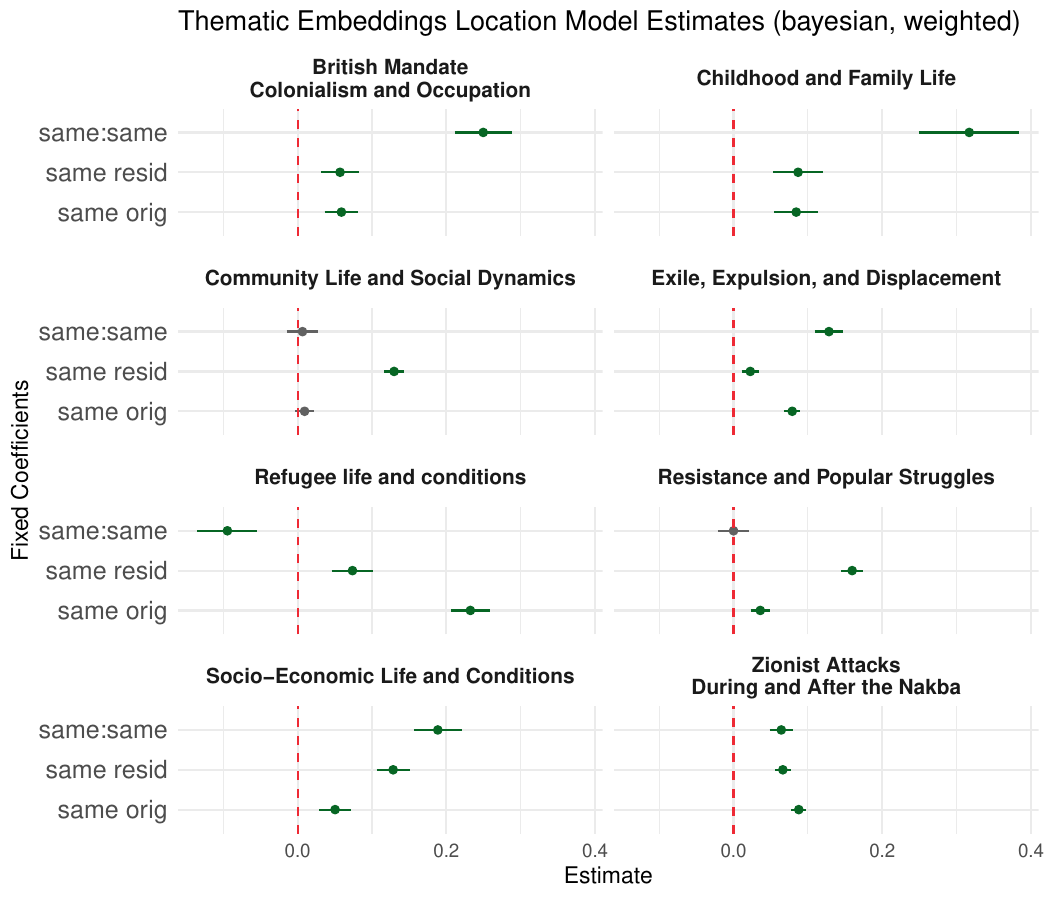}
    \caption{Bayesian model estimates with credible intervals for the models predicting similarity in themes according to location matches of the pairings of interviews.}
    \label{fig:appendix:results:tocs_location_bayes_all}
\end{figure}

\begin{figure}
    \centering
    \includegraphics[width=\linewidth]{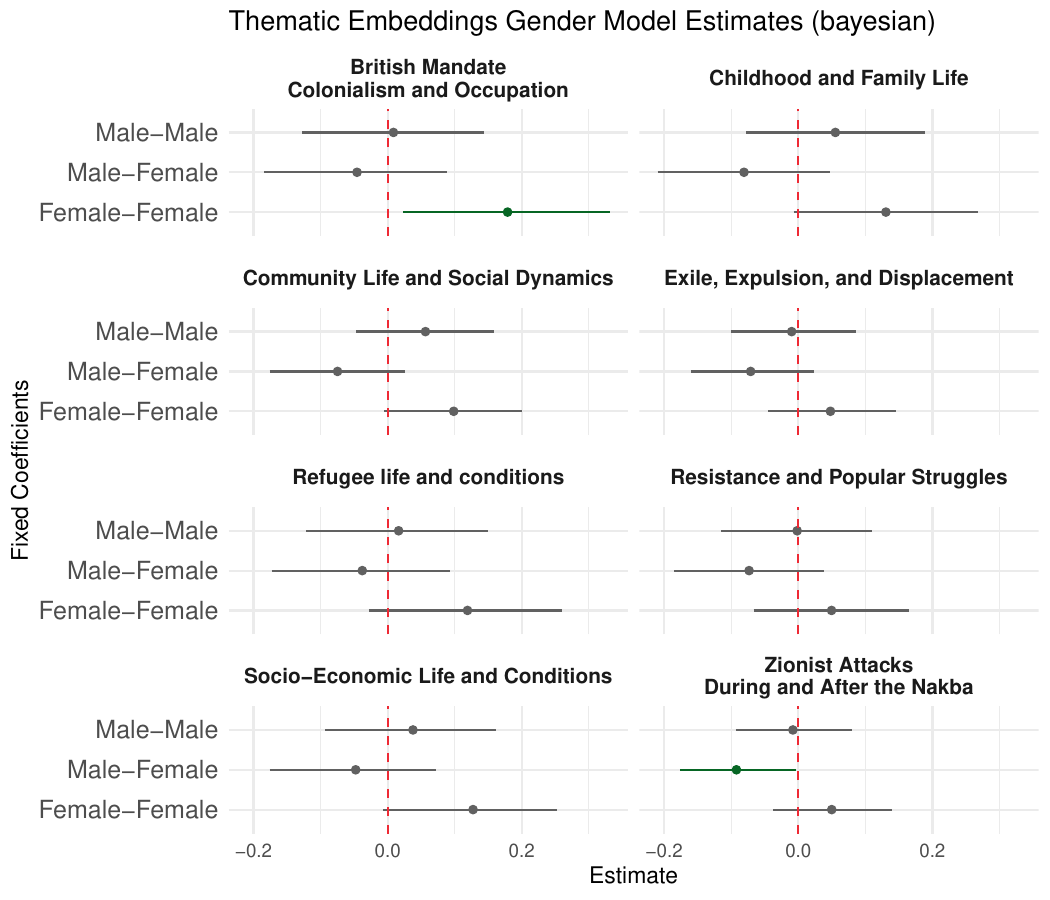}
    \caption{Bayesian model estimates with credible intervals for the models predicting similarity in themes according to gender matches of the pairings of interviews.}
    \label{fig:appendix:results:tocs_gender_bayes_all}
\end{figure}

Here we add the results of the Bayesian models on themes. Note here that although the representation we chose is the same, the intervals here mean credible intervals instead of confidence intervals. Results are for bayesian model for location cohesion analysis on \cref{fig:appendix:results:tocs_location_bayes_all} and for gender cohesion analysis in \cref{fig:appendix:results:tocs_gender_bayes_all}

\subsection{Embedding similarity insights}
POHA contains rich information about experiences in different parts of pre-1948 Palestine. Here, we present an additional table of contents-based plots.

\begin{figure*}
    \centering
    \includegraphics[width=0.9\linewidth]{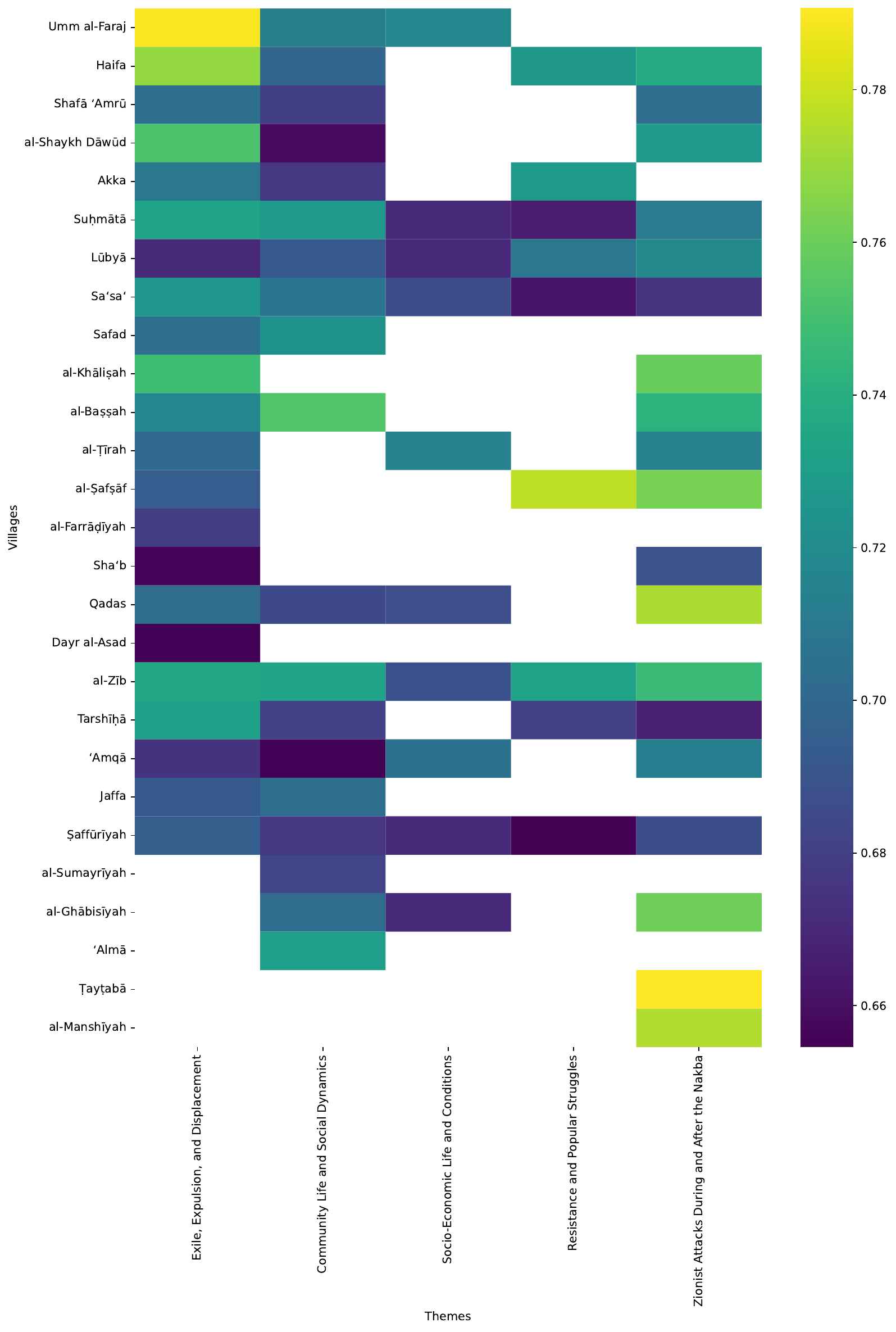}
    \caption{Average cosine similarity between the embeddings of two interview segments on the labeled theme for every place of origin in POHA. A blank rectangle represents a lack of data.}
    \label{fig:narrative_similarity_location}
\end{figure*}

We also explore the coocurrence of different themes. The theme scheme that we use engenders significant overlap among a couple of categories---namely, Zionist attacks, expulsion, and resistance.

\begin{figure*}
    \centering
    \includegraphics[width=0.9\linewidth]{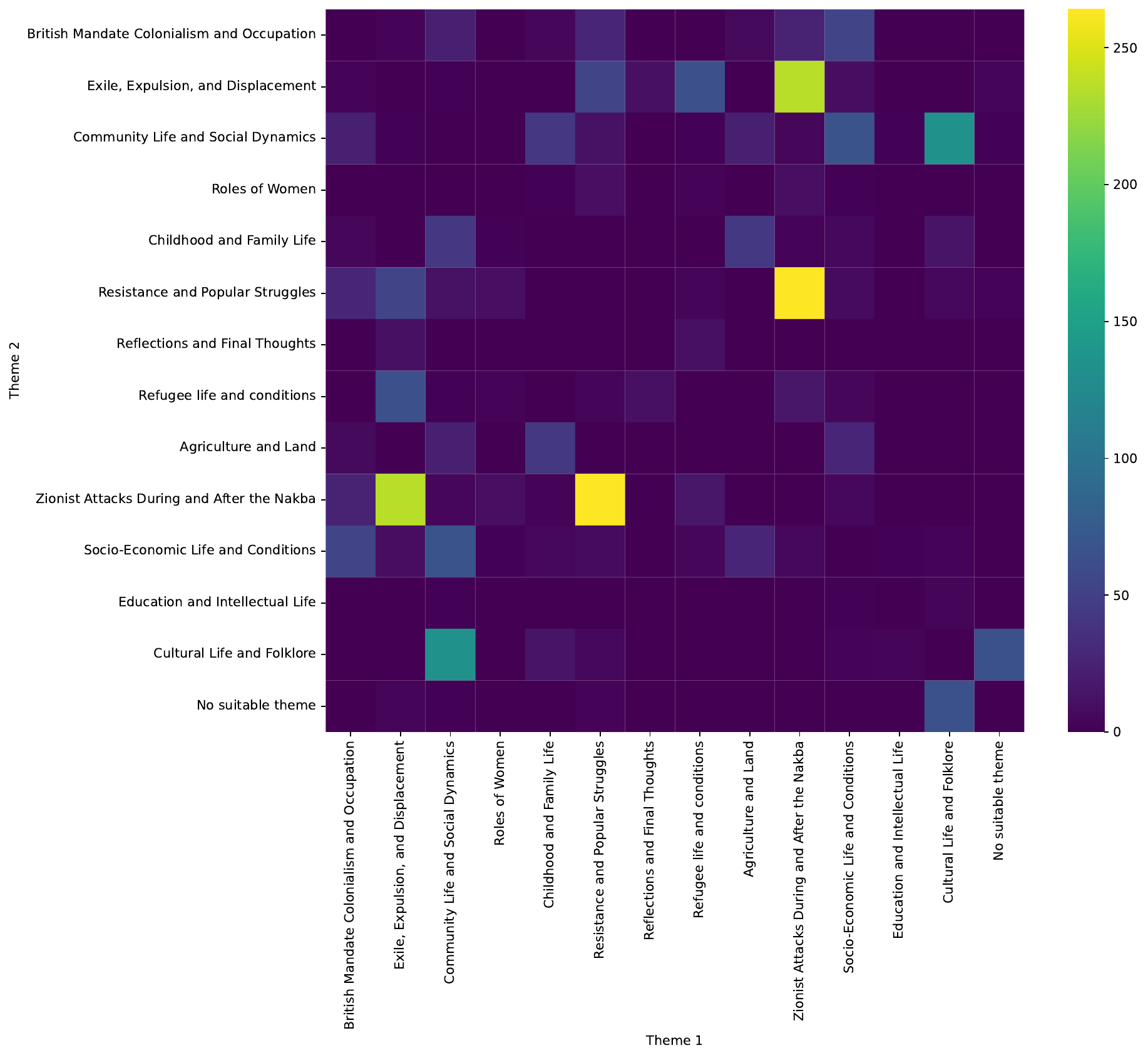}
    \caption{Cooccurrence between themes in POHA interview segments (i.e., the number of headers for which one theme and another both apply).}
    \label{fig:theme_coocurrence}
\end{figure*}

\end{document}